# Supervised Learning of Digital image restoration based on Quantization Nearest Neighbor algorithm

Md. Imran Hossain , Syed Golam Rajib

**Abstract -** In this paper, an algorithm is proposed for Image Restoration. Such algorithm is different from the traditional approaches in this area, by utilizing priors that are learned from similar images. Original images and their degraded versions by the known degradation operators are utilized for designing the Quantization. The code vectors are designed using the blurred images. For each such vector, the high frequency information obtained from the original images is also available. During restoration, the high frequency information of a given degraded image is estimated from its low frequency information based on the artificial noise. For the restoration problem, a number of techniques are designed corresponding to various versions of the blurring function. Given a noisy and blurred image, one of the techniques is chosen based on a similarity measure, therefore providing the identification of the blur. To make the restoration process computationally efficient, the Quantization Nearest Neighborhood approaches are utilized.

——————————  ◆  ——————————

## 1. INTRODUCTION

Image Restoration is a process to rebuild any image as like it was. There are different traditional approaches in this area, by utilizing priors that are learned from similar images. Original images and their degraded versions by the known degradation operators are utilized for designing restored one.

The techniques are designed using the blurred images. For each such vector, the high frequency information obtained from the original images is also available. During restoration, the high frequency information of a given degraded image is estimated from its low frequency information based on the artificial noise. For the restoration problem, a number of techniques are designed corresponding to various versions of the blurring function. Given a noisy and blurred image, one of the techniques is chosen based on a similarity measure, therefore providing the identification of the blur. To make the restoration process computationally efficient, we will apply Nearest Neighborhood approaches. In modern usual situations, a captured (recorded) image or portrait represents a blurred and noisy copy of an original one. The image degradation process may adequately modeled by a linear blur and additive noise or non- filtered process. A lot of techniques exist to determine noise or defect of image for Image Restoration. The problem typically addressed concerns characterization of the image blur caused by motion during the integration time of a photosensitive imaging medium. In this typical case the blur is usually constant throughout the image and may be caused by linear or vibratory motion [2]. Characterization of this uniform blur leads to a deconvolution type image restoration algorithm. This idea was applied in [6] to the image restoration problem, in to the super resolution problem, by designing in both cases vector quantizers (VQ), in to scene estimation [9] of super-resolution problem, and in to the blur identification problem. We have developed an algorithm to recover the parameters of the model from a noisy image.

The procedure involves alternate parameter estimation and data cleaning. We consider the estimation of the parameters of a model where the image obeying the model is not available, the corrupting innovations process being a mixture of a Gaussian process [03] and an outlier process. However the use of various stringent constraints for insufficient information in the deconvolution process limits the applications of these methods. In the presence of both blur and noise, the restoration process requires the specification of additional smoothness constraints on the solution. The objective of image restoration is to recover the high frequency information in an image that was removed by the low-pass degradation system. Therefore, in following a learning approach to restoration, the mapping between low- and high-frequency information in an image needs to be established during restoration process.

——————————————

- F.A. Author is with the Department of Electronics and Telecommunication Engineering, Daffodil International University, Dhaka, Bangladesh
- S.B. Author Jr. is with the Computer Science and Engineering Discipline , Khulna University, Bangladesh

## 2. LITERATURE REVIEW
### 2.1 Image Restoration

In computer science, image processing is any form of signal processing for which the input is an image, such as photographs or frames of video; the output of image processing can be either an image or a set of characteristics or parameters related to the image. Most image-processing techniques involve treating the image as a two-dimensional signal and applying standard signal-processing techniques to it. Image processing or Image Restoration usually refers to digital image processing, but optical and analog image processing are also possible. This article is about general techniques that apply to all of them. The *acquisition* of images (producing the input image in the first place) is referred to as imaging. Many applications such as medical imaging, radio astronomy and remote sensing observed images are sometimes degraded by distortion [7].

Distortion may arise from atmospheric turbulence, relative motion between an object and the camera and an out- of- focus camera. Restoration of degraded images is generally required for further processing or interpretation of the images. Because constraints on the degradation and the original image vary with the application, many different algorithms exist to solve the problem. In some cases, the original image, which is modeled as either a deterministic is blurred by a known function. Many different conventional approaches have been developed to compensate for blur functions when they are known. More commonly the blur function is not known. In this case the model of the blur is often assumed, for example, a linear space- invariant filter. In some applications, several blurred versions of the same original image come from different blurring channels, or several blurred images are available from but highly-correlated original images and channels, as in short-exposure image sequences, multispectral images and microwave radiometric images. Restoration of this image or images called Image Restoration.

In general, blurred image $x(n_1, n_2)$ can be modeled as the equation bellow [7]:

$$x(n_1, n_2) = T\left(\sum\sum h(n_1, n_2, l_1, l_2) s(n_1, n_2)\right) \cdot w(n_1, n_2)$$

Our work is to restore single- channel image noise. Two classes of techniques are very popular and developed. One class utilizes the stochastic models for the original models, the blur function or the noise. The unknown noise parameters are estimated from the blurred image using different approaches. The restored can be obtained from these parameters.

The field of image restoration is generally concerned with the estimation of uncorrupted images from the noisy and blurred images acquired by imaging systems [3] [4]. In this research, we are only concerned with the blur caused by diffraction- limited optics within incoherent illumination. Optical systems with these characteristics can be modeled as linear, shift-invariant operators with well defined optical transfer functions derived from the scalar diffraction theory of light as well. The inverse problem is difficult to solve due its ill-posed nature of any image. In this case, the absence of noise, the relationship between the object and image is unique in theory and exact solution is possible. Noise undermines this relationship and makes an exact solution impossible. In fact, a solution is usually defined as an element of a set of feasible solutions consistent with the linear relationship. One then seeks to incorporate prior knowledge in order to narrow the choices and produce better estimates.

### 2.2 Image Restoration

In pattern recognition, [3][7]the *k*-nearest neighbors algorithm (*k*-NN) is a method for classifying objects based on closest training examples in the feature space. *k*- NN is a type of instance-based learning, or lazy learning where the function is only approximated locally and all computation is deferred until classification. It can also be used for regression. The *k*-nearest neighbor algorithm is amongst the simplest of all machine learning algorithms.

An object is classified by a majority vote of its neighbors, with the object being assigned to the class most common amongst its *k* nearest neighbors. *k* is a positive integer, typically small. If *k* = 1, then the object is simply assigned to the class of its nearest neighbor. In binary (two class) classification problems, it is helpful to choose *k* to be an odd number as this avoids tied votes.

The same method can be used for regression, by simply assigning the property value for the object to be the average of the values of its *k* nearest neighbors. It can be useful to weight the contributions of the neighbors, so that the nearer neighbors contribute more to the average than the more distant ones.

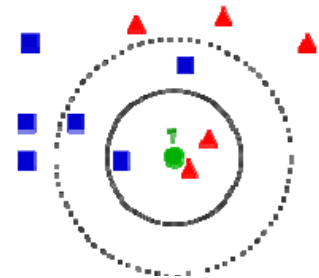

Figure 1: *k*- Nearest Neighbor

The neighbors are taken from a set of objects for which the correct classification (or, in the case of regression, the value of the property) is known. This can be thought of as the training set for the algorithm, though no explicit training step is required. In order to identify neighbors, the objects are represented by position vectors in a multidimensional feature space. It is usual to use the Euclidean distance, though other distance measures, such as the Manhattan distance could in principle be used instead. The *k*-nearest neighbor algorithm is sensitive to the local structure of the data.

These are the steps of general Nearest Neighbor algorithm:

1. Stand on an arbitrary vertex as current vertex.
2. Find out the lightest edge connecting current vertex and an unvisited vertex V.
3. Set current vertex to V.
4. Mark V as visited.
5. If all the vertices in domain are visited, then terminate.
6. Go to step 2

## 2.3 Quantization

The dictionary (*Random House*) definition of quantization is the division of a quantity into a discrete number of small parts, often assumed to be integral multiples of a common quantity. Any real number $x$ can be rounded off to the nearest integer, say $q(x)$ [05] with the resulting quantization error $e = q(x) - x$, so that the $q(x) = x + e$. More generally, we can define a quantizer as consisting [1] of a set of intervals $S = \{S_i, i \in I\}$, where the index set is $I$ ordinarily a collection of consecutive integers beginning with or , together with a set of reproduction values or points or levels $C = \{y_i, i \in I\}$, so that the overall quantizer $q$ is defined by $q(x) = y_i$ for $x \in S_i$, which can be expressed by

$$q(x) = \sum_i y_i \, 1_{S_i}(x)$$

Where the indicator function $1_{S_i}(x)$ is 1 if $x \in S$ and $0$ otherwise. For this definition to make sense we assume that $S$ is a partition of the real line. That is, the cells are disjoint and exhaustive. The general definition reduces to the rounding off.

The quality of a quantizer [02] can be measured by the goodness of the resulting reproduction in comparison to the original. One way of accomplishing this is to define a distortion measure $d(x, \hat{x})$ that quantifies cost or distortion resulting from reproducing $x$ as $\hat{x}$ and to consider the average distortion as a measure of the quality of a system [09], with smaller average distortion meaning higher quality. The most common distortion measure is the squared error--- $d(x, \hat{x}) = |x - \hat{x}|^2$ , but we shall encounter others later. In practice, the average will be a sample average when the quantizer is applied to a sequence of real data, but the theory views the data as sharing a common probability density function (pdf) $f(x)$ corresponding to a generic random variable $X$ and the average distortion becomes an expectation [08]

$$D(q) = E[d(X, q(X))] = \sum_i \int d(x, y_i) f(x) dx$$

If the distortion is measured by squared error, $D(q)$ becomes the mean squared error (MSE), a special case on which we shall mostly focus.

## 2.4 High and Low Frequency Components of an Image

Qualitatively speaking, the objective of image restoration is to recover the high frequency information in an image that was removed by the low-pass degradation system. Therefore, in following a learning approach to restoration, the mapping between low- and high-frequency information in an image needs to be established during training. During restoration the low-frequencies in the available noisy and blurred data are mapped with the use of a codebook to high frequencies, which when added to the available data provide an estimate of the original image. Decomposition of an image into low- and high-frequency components is first performed. That is, if matrix $H$ represents a low-pass operator, an image $f$ can be written as

$$f = Hf + (I - H)f = f_{(H)} + f_{(I-H)}$$

Where $I$ is the identity matrix and $f_{(H)} = Hf$ and $f_{(I-H)} = (I - H)f$ are respectively the low and the high-frequency components of $f$.

A mapping $V_H$ can therefore be defined between a low- and a high-frequency neighborhood in an image, that is

$$V_H(f_H) : f_H \rightarrow f_{(I-H)}$$

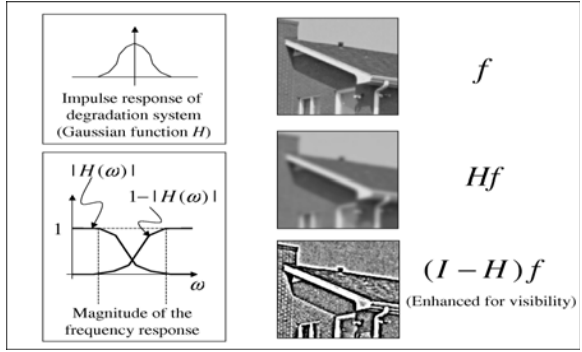

Figure 2: High and Low pass filtered image

With this mapping, it can be rewritten as follows

$$f = f_{(H)} + f_{(I-H)} \approx f_H + V_H(f_{(H)})$$

If the mapping $V_H$ is available, only the low frequency information of the image is required to get an approximation of the original one. Specifying $V_H$ is the objective of the off-line training process.

Another experiment was carried out by assuming that the blur function, which is parameterized by the radius $r$ according to the equation provided bellow.

$$h(i,j) = \begin{cases} \frac{1}{\pi r^2}, & \text{if } \sqrt{i^2+j^2} \leq r \\ 0, & \text{otherwise} \end{cases}$$

### 2.5 Blur Identification Equation

Another experiment was carried out by assuming that the blur function is a pillbox function, which is parameterized by the radius $r$ according to the equation provided bellow.

$$h(i,j) = \begin{cases} \frac{1}{\pi r^2}, & \text{if } \sqrt{i^2+j^2} \leq r \\ 0, & \text{otherwise} \end{cases}$$

The results we will obtain demonstrate that the same trend as shown in Fig. 2 and that the correct blurring parameter is identified by selecting the codeword with the minimum distortion. For both cases, the accuracy of the identification result does not seem to depend on the noise power.

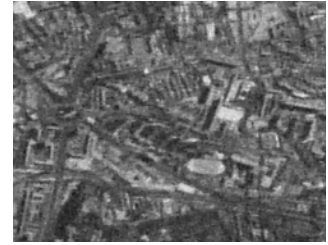

Figure 3: Noisy Image

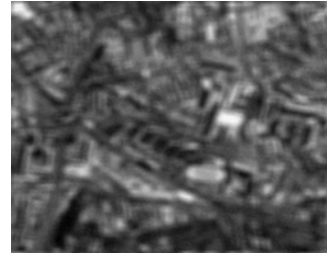

Figure 4: Degraded Image

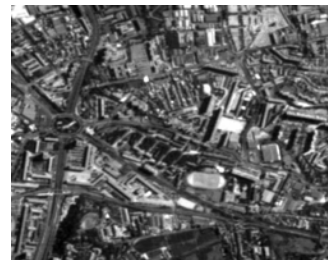

Figure 5: Original Image

### 2.6 Image Restoration Equation

Several studies have been performed on the effectiveness of the Image Restoration algorithms. In this section, we have described the experimental equation with the proposed image restoration algorithm. We will deal with images degraded by the Gaussian degradation function and additive zero-mean Gaussian noise. The Gaussian degradation function is expressed as follows:

$$h(i,j) = K \exp\left(-\frac{i^2+j^2}{2\sigma^2}\right)$$

Where $K$ is a normalizing constant ensuring that the blur is of unit volume and $\sigma^2$ is the variance.

## 3. PROPOSED QNN ALGORITHM

The approach we have described here for image restoration is extended here to include the blur identification problem. The basic additional element of the approach is the given image that is artificially blurred by blur equation. The QNN (Quantization Nearest Neighbor) algorithm adapted by the proposed algorithm is briefly described next in the following we use $T$ to represent the total number of labels and the term codeword to describe the pair of a low-frequency vector and a high-frequency scalar.

**Steps of proposed algorithm:**

Step 1:

The original image is blurred by blur function $g = M[i][j]f + n$, where g means blurred image, $f$ means original (scanned) image, $n$ means noise and $M$ means linear degradation matrix.

Step 2:

Here we will find two types of region- Nonflat Region and Flat Region. A predefined value (threshold) will be used to identify the flat and nonflat regions. This is formulated y the quantization equation $q(x) = \sum_i y_i I_{S_i}(x)$, where $x \in S$ and $0$ otherwise. For example the set $S$ contains values from $\{1,2,3,......,255\}$ for any low band image.

Step 3:

By those values or codewords from Step 2, it will make each codebook. Here Distance Transform is calculated by $d(x,f) = |x - f|^2$.

Step 4:

Applying NNN algorithm

{
Step 0: The lowest level, N=1
  Produce a Distance Image D of the same size as the corrupted image. For (r, c) in image, if l[r][c] is NOT corrupted D[r][c] =0. Else D[r][c] = ∞ (typically represented by 255 in unsigned char).
If N> 1
{
Step 1:

The most general NNN restoration Algorithm, is for neighbourhood classification containing at least the N=n >1 closest neighbours.In this case the algorithm proceeds to Step 2.
Step 3:
For the pixel at (r.c) at a non-zero distance d. determine the set S(d, r.c) of good pixels at distance d from (r.c). Then determine S(d+l. r.c). S(d+2. r,c) ... until at least n good nearest neighbour pixels have been determined
Step 4:
Generate the restored image R: If D[r][c]=O, R[r][c] =l[r][c],
else R[r][c] = Classification Algorithm applied to the scalar valued gray scale or vector-valued coloured pixels in the sets
{S(d.r.c), S(d+l.r.c),....}.

}

Step 2:

Apply the DT of choice to determine the distance of every corrupt pixel from its nearest good pixel.

Step 3:

For the pixel at (r,c) at a non-zero distance determine the set S(r.c) of good pixels at distance d from (r.c).

Step 4:

Generate the restored image R: If D[r][c] = 0, R[r][c] = l[r][c] else R[r][c] = lazy median of S(r,c); for gray-scale images. the actual median if it exists, with random choice fixing selection otherwise. For the restoration of the indexed color.

}
##End of Restoration

## 4. EXPERIMENTAL RESULTS

Several experiments have been performed in studying the effectiveness of the proposed algorithms. In this section, we first describe some of the experimental results obtained with the proposed *Image Restoration Algorithm* and then some of the experimental results obtained with the given *Blur Identification Equation* in section 2.5.

In our experiments, we dealt with images degraded by the Gaussian degradation function and additive zero-mean Gaussian noise. The Gaussian degradation function is expressed as follows:

$$h(i,j) = K \exp\left(-\frac{i^2 + j^2}{2\sigma^2}\right)$$

where $K$ is a normalizing constant ensuring that the blur is of unit volume and $\sigma^2$ is the variance. We experimented with values of $\sigma^2$ equal to 1.5 and 3.5 and two levels of noise resulting in values of blurred SNR (BSNR) of 20 and 10 dB. The BSNR in dB for an $I \times J$ image is defined by

$$BSNR = 10 \times \log_{10}\left\{\frac{1}{IJ} \frac{1}{\sigma_n^2} \sum_{i,j}[Df - E\{g\}]^2\right\}$$

Where $D$ and $f$ are respectively the blurring function and the original image, $E\{g\}$ is the expected value of the degraded image and $\sigma_n^2$ the variance of the noise.

For comparison, we present restoration results obtained by the Constrained Least Squares (CLS) filter. This is a widely used linear restoration technique [1]–[3]. A 3×3 Laplacian filter is used to implement the CLS filter and the regularization parameter is chosen as the reciprocal of the BSNR of the observed image. The choice of 1/BSNR can be justified through the set-theoretic formulation of the problem, where the loose bounds on the constraints we try to satisfy are the noise variance and the high frequency signal variance.

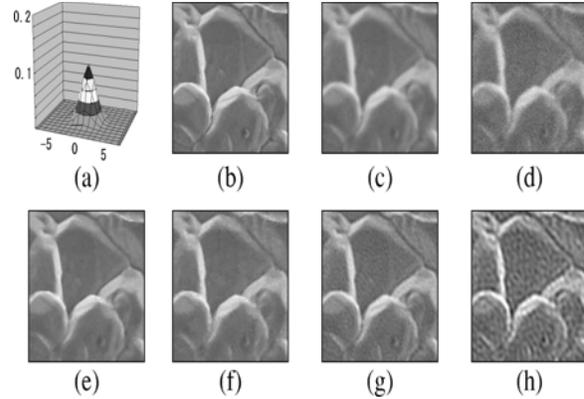

Figure 6: (a) Gaussian Blur Function (variance = 1:5); (b) Original image; (c) Degraded image, BSNR = 20 dB; (d) Degraded image, BSNR = 10 dB; (e) Restored image of (c) by proposed algorithm, (ISNR = 2:91 dB); (f) Restored image of (d) by proposed algorithm (ISNR = 3:06 dB); (g) Restored image of (c) by CLS algorithm (regularization parameter = 0:05, ISNR = 2:46 dB); (h) Restored image of (d) by CLS algorithm (regularization parameter = 0:1, ISNR = 0:87 dB).

The simulated degraded versions of these images are calculated by the same degradation system by which the given image is corrupted and these pairs are used for the design of the codebook. Fig. 7(b) shows the 32 representative vectors calculated by the LBG algorithm for the Gaussian blur function with variance equal to 1.5 and BSNR = 20 dB. Each block image is of size 7×7 pixels. The contrast of these block images is enhanced for viewing purposes. These representative blocks consist of various edges of different directions, amplitudes, and frequency. Vertical edges are over-represented in this set since they are dominant in the prototype images.

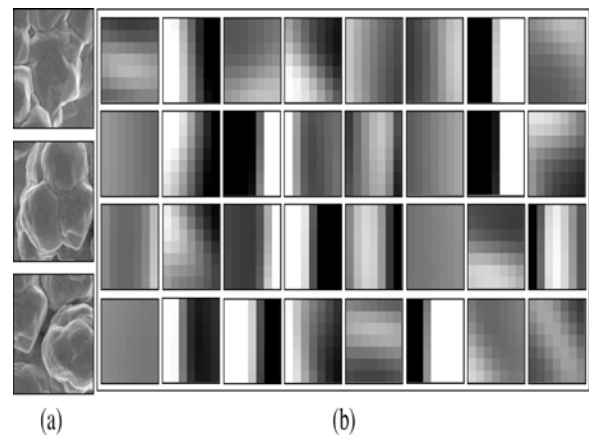

Figure 7: (a) Images used for the codebook design.(b)

The 32 representative vectors calculated by the LBG A similar set of restoration results is shown in Fig.

8 when a Gaussian blur of variance equal to 3.5 is used (shown in Fig. 8(a)). For the case, most of the edges are recovered by the proposed method (Fig. 8(e)). On the contrary, the results by the CLS filter (Fig. 8(g)) suffers from noise amplification. For the case, noise amplification is strongly visible in the result by the CLS algorithm (Fig. 8(h)). Although, the result by the proposed method does not suffer from noise amplification, parts of the straight edges are not well restored as expected. The ISNR values of the results obtained by the proposed method are 3.04 and 2.85 dB for and 10 dB, respectively. These values are similar or higher than those obtained with the CLS algorithm (3.18 dB and 1.47 dB), respectively.

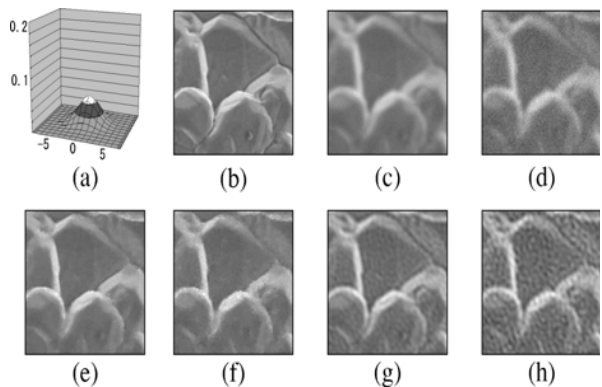

Figure 8: (a) Gaussian Blur Function (variance = 3:5). (b) Original image. (c) Degraded image, BSNR = 20 dB. (d) Degraded image, BSNR = 10 dB. (e) Restored image of (c) by proposed algorithm (ISNR = 3:04 dB). (f) Restored image of (d) by proposed algorithm (ISNR = 2:85 dB). (g) Restored image of (c) by CLS algorithm (regularization parameter = 0:05, ISNR = 3:18 dB). (h) Restored image of (d) by CLS algorithm (regularization parameter = 0:1, ISNR = 1:47 dB).

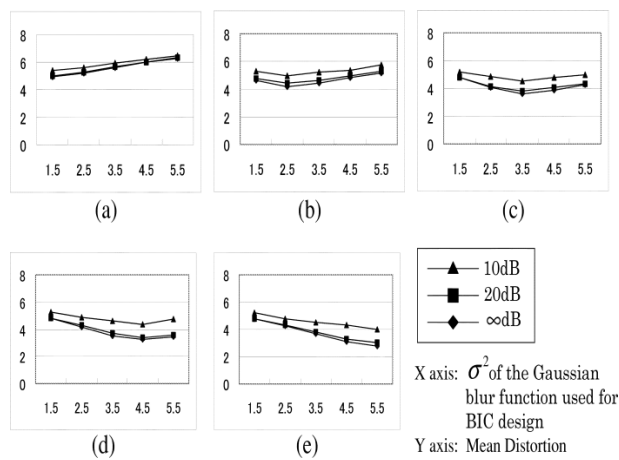

Figure 10: Mean distortion $d_i$ between a given degraded image and the blur identification codebook. The variance $\sigma^2$ of the Gaussian blur function resulting in the degraded data is equal to: (a) 1.5, (b) 2.5, (c) 3.5, (d) 4.5, and (e) 5.5.

Here in Fig 10 shows the CLS algorithm produces good results for the fine texture region area, although, noise amplification in the flat regions such as, the cheek and the forehead, is strongly visible. This shows that the result by the proposed method may look perceptually better than the CLS algorithm even though the ISNR of the restored image by the proposed method is equal to 0.95 dB and it is lower than that of the CLS algorithm, which is equal to 1.93 dB. Fig. 10 shows the relationship between the calculated mean distortion value $d_i$ in Fig. 5, for a given degraded image and the blur identification codebook for 3 noise levels. The graphs (a)–(e) correspond to the cases where the blur parameter of a given blurred image is equal to $\sigma^2 = 1.5$, 2.5, 3.5, 4.5 and 5.5 respectively. In each graph, the $Y$- axis shows the distortion between the given image and the BIC, and the $X$ - axis shows the $\sigma^2$ of the blurred image used for the blur identification codebook (BIC) design. We can see that in all cases the larger the noise power, the larger the distortion. The plots in Fig. 11 demonstrate the existence of a minimum point. This means that we can identify the correct parameter by selecting the codebook with the minimum distortion.

## 5. CONCLUSIONS

In this paper, we developed a Quantization based image restoration algorithm. In the image restoration algorithm, the mapping between high frequency information of the original images and low frequency information of the corresponding degraded ones is established and stored in the codebooks, using prototype images. Experimental results demonstrate that the proposed image restoration algorithm restores the sharp edges in the image without suffering from noise amplification. The results also show that the proposed blind restoration algorithm correctly identifies the blurring function. The mapping between high and low frequencies can be implemented not only by VQ but also by other nonlinear estimation techniques, such as ANN (Artificial Neural Network) and SVN (Support Vector Machine). These techniques may improve the performance of the restoration algorithm in certain applications.

**Md.Imran Hossain** received his B.Sc. degree in Computer Science and Engineering from Jahangirnagar University, Savar, Dhaka, Bangladesh in 2004 and M.Sc. degree in Telecommunication Engineering from Blekinge Institute of Technology (BTH), Sweden in 2008. Now he is serving as a Lecturer in the department of Electronics and Telecommunication Engineering at Daffodil International University, Dhaka, Bangladesh. His field of research interest includes wireless and mobile communication, Mobile Ad-hoc Network (MANET).

**Syed Golam Rajib** is a student of B.Sc. in Computer Science and Engineering from Khulna University, Khulna, Bangladesh . His research interest includes Microwave Engineering, Pattern recognition and Image processing.